\documentclass[letterpaper, 10 pt, conference]{ieeeconf}  
\IEEEoverridecommandlockouts                              
\usepackage{graphics} 
\usepackage{epsfig} 
\usepackage{mathptmx} 
\usepackage{times} 
\usepackage{amsmath} 
\usepackage{amssymb}  
\usepackage{hyperref}

\usepackage{multirow}
\usepackage{subfigure}
\usepackage{booktabs}
\usepackage{array, caption, threeparttable}
\usepackage[font=small]{caption}

\usepackage{times}
\usepackage{epsfig}
\usepackage{graphicx}
\usepackage{float}
\usepackage{wrapfig}
\usepackage{amsmath,amssymb,amsthm}
\usepackage{algorithm,algorithmicx,algpseudocode}
\usepackage{bm,xspace}
\usepackage{comment}
\usepackage{verbatim}
\usepackage{multirow}
\usepackage{balance}
\usepackage{url}
\usepackage{booktabs}
\usepackage{etoolbox,siunitx}
\usepackage{calc}
\usepackage{pifont,hologo}
\usepackage[usenames, dvipsnames]{color}
\usepackage{nicefrac}

\newcommand{\ra}[1]{\renewcommand{\arraystretch}{#1}}

\usepackage[super]{nth}
\usepackage{nicefrac}
\robustify\bfseries

\def\tt{\mathbf{t}}

\DeclareMathSymbol{@}{\mathord}{letters}{"3B}

\newcommand\mypara[1]{\vspace{1mm}\noindent\textbf{#1}}

\def\latex/{\LaTeX}
\def\bibtex/{\hologo{BibTeX}}

\title{\LARGE \bf
Stereo Waterdrop Removal with Row-wise Dilated Attention
}

\author{Zifan Shi, Na Fan, Dit-Yan Yeung, and Qifeng Chen
\thanks{Zifan Shi (zshiaj@connect.ust.hk), Na Fan (nfanaa@connect.ust.hk), Dit-Yan Yeung (dyyeung@cse.ust.hk) and  Qifeng Chen (cqf@ust.hk) are with the Department of Computer Science and Engineering, HKUST.}
}

\begin{document}

\maketitle
\thispagestyle{empty}
\pagestyle{empty}
\begin{abstract}
 Existing vision systems for autonomous driving or robots are sensitive to waterdrops adhered to windows or camera lenses. Most recent waterdrop removal approaches take a single image as input and often fail to recover the missing content behind waterdrops faithfully. Thus, we propose a learning-based model for waterdrop removal with stereo images. To better detect and remove waterdrops from stereo images, we propose a novel row-wise dilated attention module to enlarge attention's receptive field for effective information propagation between the two stereo images. In addition, we propose an attention consistency loss between the ground-truth disparity map and attention scores to enhance the left-right consistency in stereo images. Because of related datasets' unavailability, we collect a real-world dataset that contains stereo images with and without waterdrops. Extensive experiments on our dataset suggest that our model outperforms state-of-the-art methods both quantitatively and qualitatively. Our source code and the stereo waterdrop dataset are available at \href{https://github.com/VivianSZF/Stereo-Waterdrop-Removal}{https://github.com/VivianSZF/Stereo-Waterdrop-Removal}


\end{abstract}
\section{INTRODUCTION}

Waterdrops adhered to windows or camera lenses can occlude and deform part of a captured image so that the perception systems in autonomous driving, robots, drones, and surveillance systems may not perform properly~\cite{halimeh2009raindrop,kurihata2005rainy}. The inconvenience brought by unwanted waterdrops promotes the demand for robust waterdrop removal in robotic applications, especially on rainy days.

Recent learning-based waterdrop removal methods are mainly designed for a single image~\cite{eigen2013restoring,li2020all,qian2018attentive,quan2019deep}, multi-images~\cite{liu2020learning}, and videos~\cite{alletto2019adherent,guo2018on,wu2012raindrop}. Single-image methods localize the waterdrops by estimating a waterdrop map or utilizing the physical shape priors. However, they suffer from noisy ground-truth difference map and not always obvious physical priors, resulting in the difficulty in detecting some types of waterdrops, e.g., those blend into the background scene indiscriminately or are in small sizes. Besides, they rely on prior knowledge or surrounding pixels to infer the missing details that are unreliable. The ambiguity in single-image methods can be addressed with multi-images or videos. Multi-image methods utilize the multi-view information captured by multiple cameras to remove waterdrops. Video-based methods require the motion of waterdrops, scenes, or cameras during the capturing period and aggregate features from multi-frame unobstructed areas. Though the multi-image and video-based methods reduce the ambiguity, they are not user-friendly as multiple cameras are not easily accessible, and taking a video is time-consuming.

To address the aforementioned issues, we study a novel method using stereo image pairs for robust waterdrop removal. The stereo image pair $I_l$ and $I_r$ with waterdrops can be formulated as~\cite{you2013adherent}:
\begin{align}
    I_l &= (1-T_l)\odot O^*_l + T_l\odot R_l, \\
    I_r &= (1-T_r)\odot O^*_r + T_r\odot R_r,
\end{align}
where $\odot$ means element-wise multiplication. $O^*_l, O^*_r\in\mathbb{R}^{H\times W\times C}$ are waterdrop-free background images from the left and right views, respectively. $R_l, R_r\in\mathbb{R}^{H\times W\times C}$ denote the waterdrops. They can be different regarding the spatial distributions, sizes, and shapes of the waterdrops. $T_l, T_r\in\mathbb{R}^{H\times W\times C}$ denote the transparency matrix, where each element represents how much information of the background scene is covered by the waterdrop in the corresponding pixel. Different transparency may make the waterdrops with similar aforementioned characteristics have completely different appearances due to illumination and background scenes from different perspectives. To detect and remove the waterdrops robustly, we encode the stereo image pair to multi-scale features and conduct detection and removal on each scale.

The key challenge of stereo waterdrop removal is how to utilize and maintain the left-right consistency in stereo images. Recent CNN-based stereo matching methods~\cite{chang2018pyramid,xu2020aanet,yang2019hierarchical,zhang2019ga} use 3D or 4D cost volumes to model the consistency. Cost volumes can be incorporated into the waterdrop removal architectures, but they suffer from high computational cost and the ambiguity in construction brought by corrupted regions. A more flexible choice is parallax attention~\cite{wang2019learning}, inspired by non-local attention~\cite{wang2018non}. They use 1x1 convolution kernels to extract the query and the key, and thus the receptive fields in these attention modules are relatively small compared to the size of large waterdrops, especially in low-level features. Therefore, not much valid information is provided for the query to find the corresponding key in the other feature map.

In this paper, we propose a row-wise dilated attention module (RDA) to enlarge the receptive field for effective left-right information propagation, inspired by \cite{chen2017rethinking}. In each feature level, the query and the key are extracted with several dilated convolutions to aggregate the reliable information from larger areas for left-right matching, while the key is generated with the typical 1x1 convolutions to keep sharp features. Attention is operated within a few rows with the epipolar constraint in calibrated stereo images. To further enhance the consistency, an attention consistency loss is proposed for evaluating the consistency between ground-truth disparity maps and attention scores in each level. We have conducted various controlled experiments to demonstrate the effectiveness of the proposed method.

Since there is no existing dataset for stereo waterdrop removal, we collect a stereo waterdrop dataset with 837 stereo image pairs from 129 indoor and outdoor scenes. Different from previous real-world waterdrop removal dataset~\cite{qian2018attentive}, our dataset includes more diverse waterdrops, e.g., mist-like waterdrops that are commonly seen on the window. Extensive experiments on our dataset demonstrate that our proposed method outperforms state-of-the-art methods both quantitatively and qualitatively, generating detail-preserving and visually-pleasing images similar to the ground-truth images.

Our contributions can be summarized as follows:
\begin{itemize}
    \item We are the first to study the stereo waterdrop removal problem with learning-based approaches. The proposed method outperforms state-of-the-art methods and related baselines.
    \item We design a new row-wise dilated attention module, which can capture the correspondence in stereo images more robustly with a larger receptive field. Moreover, an attention consistency loss is proposed to enhance the left-right consistency in the feature space.
    \item We propose a new real dataset for waterdrop removal with 837 stereo image pairs from 129 indoor and outdoor scenes. The dataset is of similar size to the previous real-world dataset but contains more diverse waterdrop images.
\end{itemize}
\section{Related Work}

\subsection{Waterdrop Removal}
\mypara{Stereo and multi-images.}
Kuramoto et al.~\cite{kuramoto2002removal1} first try with two cameras but only succeed in images with a small number of waterdrops. They then propose using three cameras and the majority decision to detect waterdrops~\cite{kuramoto2002removal2}.
Based on this framework, Yamashita et al.~\cite{yamashita2004removal,yamashita2003virtual} improve waterdrop detection by adding more judgment details. To further deal with close scenes or dynamic waterdrops, they turn back to the widely used stereo cameras and try to interpolate the waterdrop covered region with the disparities~\cite{yamashita2005removal}. These traditional methods do not leverage the power of deep learning for feature learning.
Recently, Liu et al.~\cite{liu2020learning} propose a general framework to remove obstructions such as reflections and waterdrops with five images from different viewpoints, leveraging motion differences between the scene and obstruction layers to recover both. However, they will introduce noises from other views to the query view. 

\mypara{Videos.}
With videos collected by a monocular camera in a moving vehicle, 
Roser and Geiger~\cite{roser2009video} detect waterdrops by comparison with rendered waterdrops.
Wu et al.~\cite{wu2012raindrop} focus on saliency in color, texture, and shape changes.
Webster and Breckon~\cite{webster2015improved} extend the waterdrop feature descriptor and isolate it from the scene. 
Guo et al.~\cite{guo2018on} follow up, and propose varying region proposal strategies. 
You et al.~\cite{you2013adherent,you2015adherent} explore local motion and intensity derivatives of waterdrops, and restore partially and completely occluded areas in different ways.
You et al.~\cite{you2014raindrop} also study the motion and appearance features of raindrops along trajectories over frames and try to preserve motion consistency.
Alletto et al.~\cite{alletto2019adherent} add motion consistency to their self-supervised single-image raindrop detection method and use a generative adversarial network for image restoration, bringing the deep networks to video methods.

\mypara{Single image.}
The earliest single-image learning-based method by Eigen et al.~\cite{eigen2013restoring} attempts to train a shallow CNN on real data. However, the output is blurry, and it is limited in waterdrops of small sizes. 
Qian et al.~\cite{qian2018attentive} propose an attentive generative adversarial network focusing on both the raindrops and the surroundings and force the local consistency of the obstructed area. 
However, the estimated waterdrop masks in their method are inaccurate due to imperfect alignment.
Quan et al.~\cite{quan2019deep} introduce shape-driven attention for waterdrop detection following the physical properties such as convexness and closedness, which in our observation, may be violated by the wind.
Hao et al.~\cite{hao2019learning} contribute a large synthetic dataset considering various physical properties of waterdrops.
Li et al.~\cite{li2020all} utilize the physical formation participles and introduce a general framework to handle various bad-weather degradations, including waterdrops.

\subsection{Stereo Image Processing}
Stereo images provide dual-view information, and have benefited many single-image processing problems to reduce the ambiguity, e.g., image super-resolution~\cite{jeon2018enhancing, wang2019learning,xie2020non,yan2020disparity, ying2020stereo}, image deraining~\cite{zhang2020stereoderain}, image dehazing~\cite{pang2020bidnet}, and image deblurring~\cite{yan2020disparity,zhou2019davanet}. Besides, AR and VR have unlocked new potential applications of stereo image processing such as style transfer~\cite{chen2018stereoscopic, gong2018neural}. The key point of stereo image processing is to ensure consistency in stereo images. Zhang et al.~\cite{zhang2020stereoderain} fuse the feature volumes from two views through simple concatenation and ResBlock refinement for stereo image deraining. Some works~\cite{chen2018stereoscopic,gong2018neural,ying2020stereo,zhou2019davanet} estimate a disparity map through a sub-network and use it to warp the features from one view to the other, followed by concatenation or addition for blending the features. Yan et al.~\cite{yan2020disparity} adopt cost volumes to learn the correspondences and fuse two-view features with the encoded disparity features. Inspired by self-attention mechanisms~\cite{vaswani2017attention, wang2018non}, several works~\cite{pang2020bidnet,wang2019learning,ying2020stereo} utilize cross-attention or parallax-attention for stereo image processing, where single-row attention is adopted to capture the dependency, and 1x1 convolution kernels are used to extract the query, key, and value. Though these methods achieve state-of-the-art performance in their domains, the way to maintain the consistency cannot be directly transferred to the waterdrop removal task because of the corruption of the image and the limited receptive fields of the methods.



\begin{figure*}[!t]
\centering
\includegraphics[width=1\linewidth]{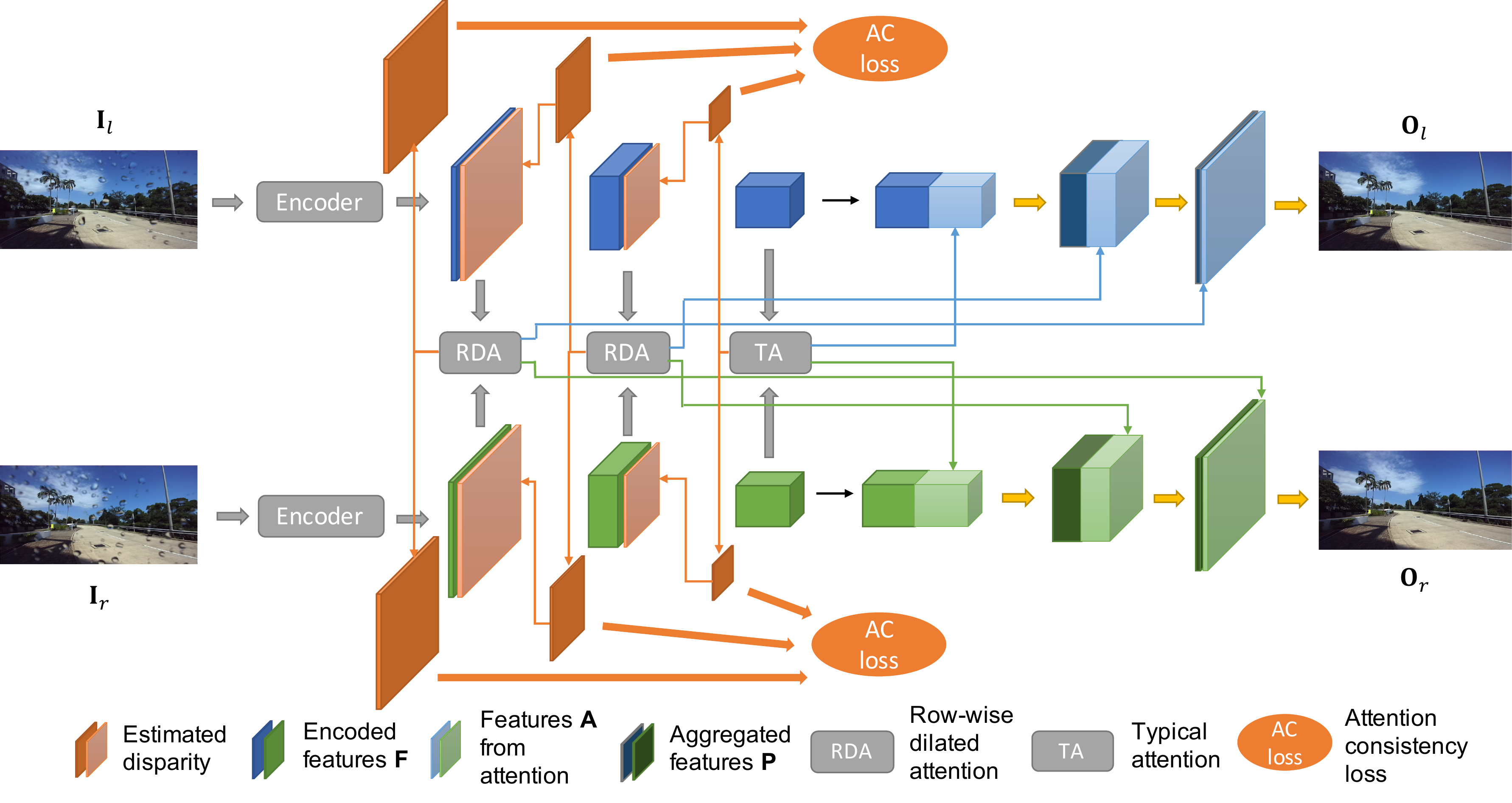}
\caption{The architecture of the proposed method for stereo waterdrop removal. The stereo image pair first goes through an encoder to generate three-level feature maps. Then the attention block operates from the highest level to the lowest level between the feature maps of the left and right images, and outputs the refined features and the estimated disparity maps in each level. The estimated disparity maps are upsampled and concatenated to the feature maps in the next lower level. Finally, the feature maps in the highest level are gradually aggregated with the refined features in each level to generate the output pair without waterdrops. Attention consistency loss is calculated according to the disparity maps estimated in three levels.}
\label{fig:overall}
\vspace{-2mm}
\end{figure*}
\section{Method}
The overall architecture of our proposed method is shown in Figure~\ref{fig:overall}. The network takes a calibrated stereo image pair $\mathbf{I}_l$ and $\mathbf{I}_r$ as input and outputs two clean images $\mathbf{O}_l$ and $\mathbf{O}_r$. Each input image is first fed into a pretrained encoder to generate a set of feature maps with different resolutions. In our case, we select ResNet-50~\cite{he2016deep} as the encoder, and choose the features maps from layers `$conv2\_3$', `$conv3\_4$', and `$conv4\_6$' to construct the three-level feature sets $\{\mathbf{F}_{l,i}\}$ and $\{\mathbf{F}_{r,i}\}$ for the left and right images. $i$ is the index for the level, ranging from 1 to 3.

\subsection{Row-wise Dilated Attention}

\begin{figure*}
\begin{tabular}{@{\hspace{5mm}}c@{\hspace{1mm}}c@{\hspace{1mm}}c@{}}
\includegraphics[width=.6\linewidth]{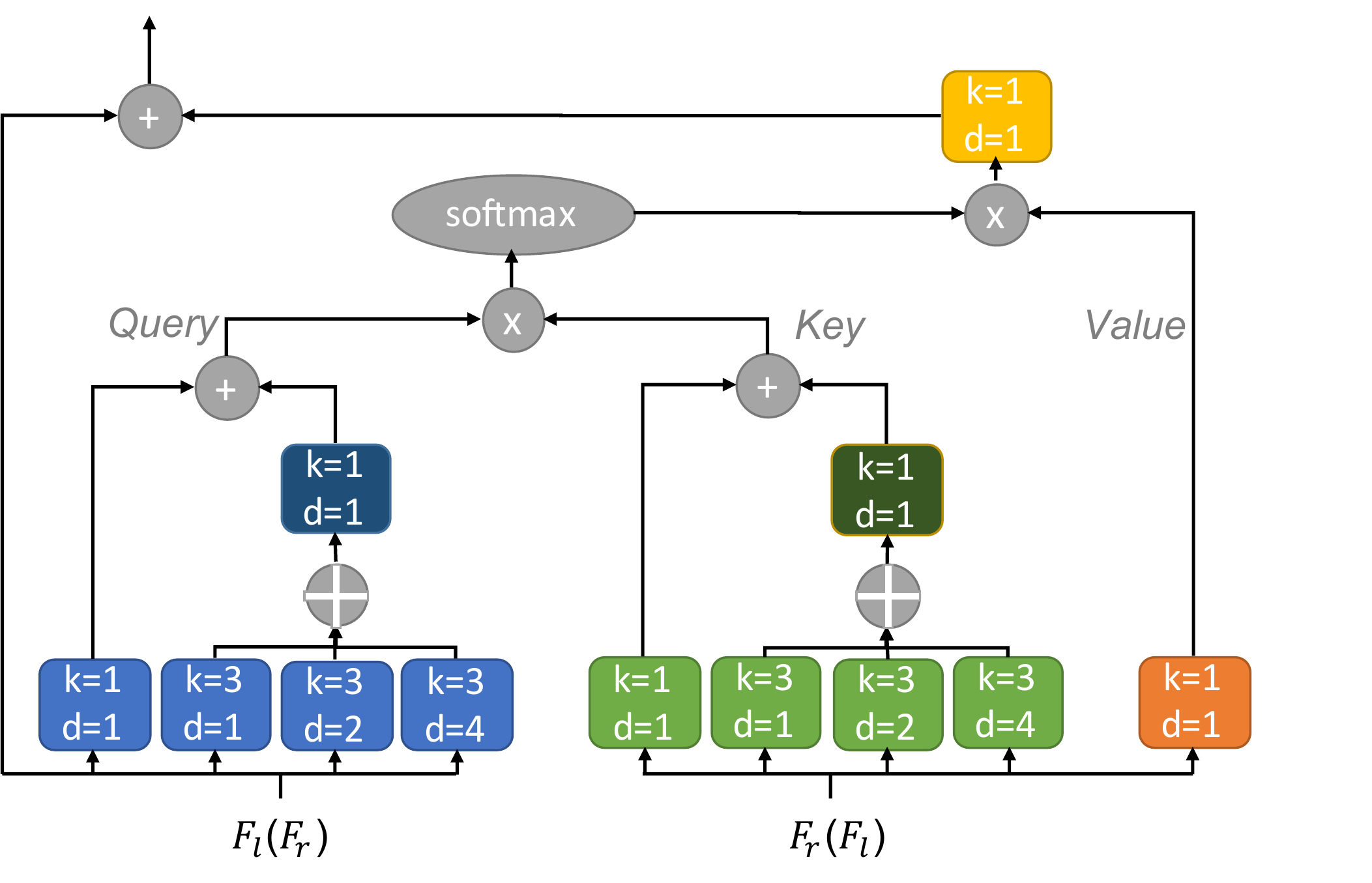} &&
\includegraphics[width=.29\linewidth]{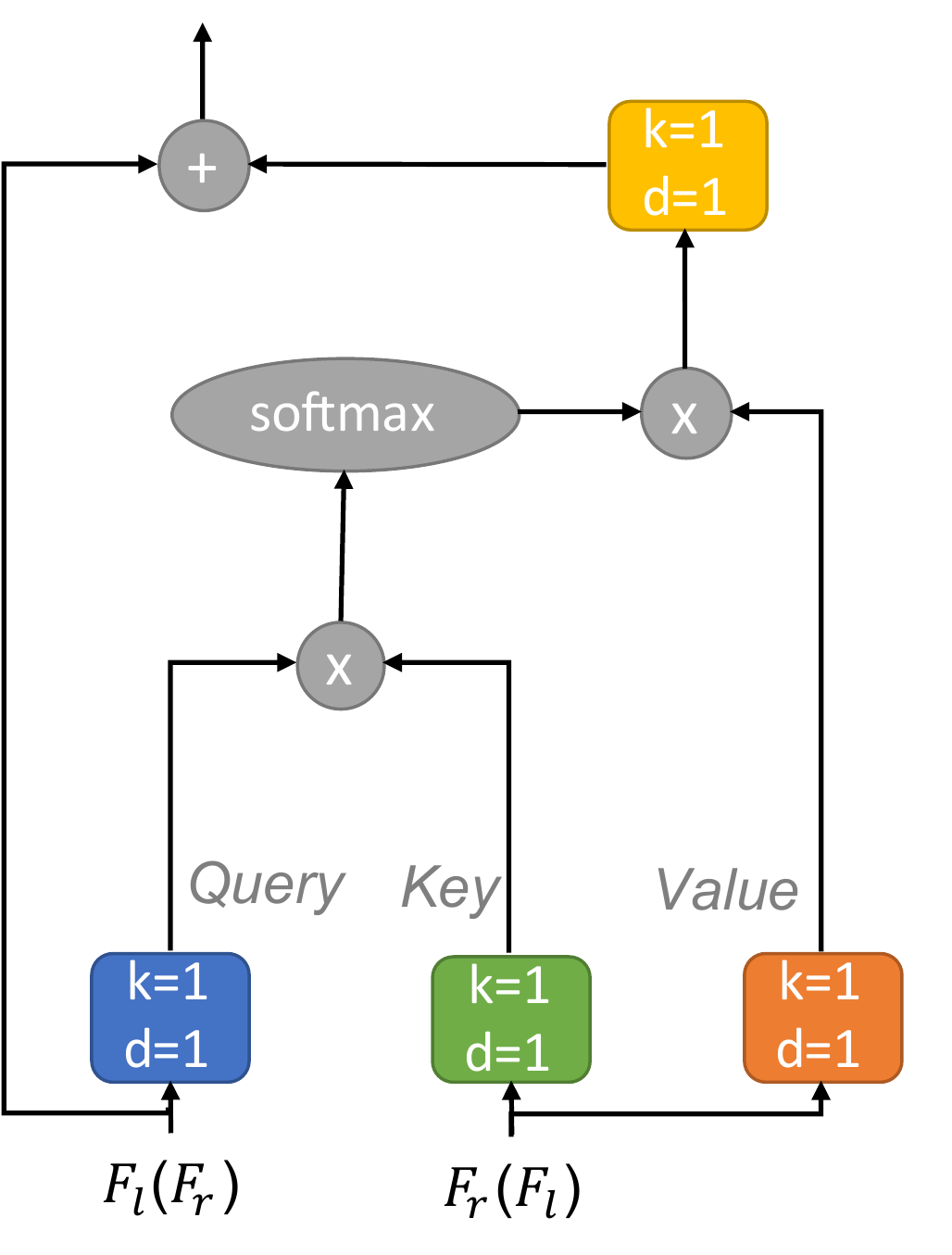}\\
\small (A) Row-wise dilated attention && \small (B) Typical non-local attention \\
\end{tabular}
\vspace{-2mm}
\caption{Two types of attention blocks. k represents the kernel size and d denotes the dilation.}
\vspace*{-3mm}
\label{fig:att}
\vspace{0mm}
\end{figure*}

The disparity in a stereo image pair makes it necessary to find the corresponding features among each other. Attention mechanism~\cite{vaswani2017attention} is a common practice for capturing global correspondence in stereo images. However, waterdrops in the images block or deform the texture information that can be used to directly locate the corresponding points in the other image. A small receptive field can be fulfilled by waterdrops, and no valid texture information is available then. Therefore, typical attention with 1x1 convolution kernels shown in Figure~\ref{fig:att} (B) fails to find the correspondence accurately, especially for low-level features. We address this issue by introducing the row-wise dilated attention (RDA), which is shown in Figure~\ref{fig:att} (A). For simplicity, we only illustrate the RDA with the left image as the query and the right image as the reference in the following. We also omit the level index $i$. The RDA with the right image as the query and the left image as the reference is the symmetric process. Note that the convolution kernels to extract the query, key, and value are shared at each level for both cases.

For feature maps $\mathbf{F}_{l}, \mathbf{F}_{r}\in \mathbb{R}^{H\times W\times C}$ in each level, we generate the query $\mathbf{Q}_{l}$ and the key $\mathbf{K}_{r}$ with four types of convolution kernels: 1x1 convolution kernels to aggregate local features, and 3x3 convolution kernels with three different dilation factors 1, 2, and 4 to enlarge the receptive field. The outputs of dilated convolutions are concatenated and fed into the 1x1 convolution kernels to distillate the information. Therefore, enough valid texture information can be captured in the query and the key for calculating the similarities between different feature grids:
\begin{align}
    \mathbf{Q}_{l} &= \mathbf{F}_{l}*_1 \mathbf{k}_{1} + (\bigoplus_{j=1,2,4}\mathbf{F}_{l} *_j \mathbf{k}^{'}_{1})*_1 \mathbf{k}_{2},\\
    \mathbf{K}_{r} &= \mathbf{F}_{r}*_1 \mathbf{k}_{3} + (\bigoplus_{j=1,2,4}\mathbf{F}_{r} *_j \mathbf{k}^{'}_{2})*_1 \mathbf{k}_{4},
\end{align}

where $\mathbf{k}_{1}, \mathbf{k}_{2}, \mathbf{k}_{3}$, and $\mathbf{k}_{4}$ denote 1x1 convolution kernels, while $\mathbf{k}^{'}_{1}$ and $\mathbf{k}^{'}_{2}$ denote 3x3 convolution kernels. The operator $*_d$ represents convolution with dilation d. $\bigoplus$ means the concatenation of feature maps along the channel dimension. Dilated convolutions would bring noisy features for the value, and thus the value $\mathbf{V}_{r}$ is extracted with 1x1 convolution kernels only to keep the useful features:

\begin{align}
   \mathbf{V}_{r} = \mathbf{F}_{r} *_1 \mathbf{k}_{5}.
\end{align}
The output of the RDA module is thus formulated as:
\begin{align}
    \mathbf{A}_{l} = \mathbf{F}_{l} + (softmax(\mathbf{Q}_{l}^T\mathbf{K}_{r})\mathbf{V}_{r})*_1 \mathbf{k}_{6},
\end{align}
where $\mathbf{k}_6$ denotes 1x1 convolution kernels to match the number of channels with that of $\mathbf{F}_l$.

Since stereo images are calibrated to ensure that objects in the same horizontal row are consistent, there is no need to calculate the attention in the whole images. Considering that the matching points in two images may be both obstructed by waterdrops, where we need the surroundings to infer the corrupted area, the attention is conducted every three rows with stride 2.

\subsection{Feature Aggregation}

Given the feature maps $\{\mathbf{A}_{l,i}\}$ and $\{\mathbf{A}_{r,i}\}$ obtained from RDA module, the original feature maps are gradually refined from high levels to low levels, using residual blocks and skip connections. For simplicity, we take the left image as an example:
\begin{align}
    \mathbf{P}_{i} = \phi_2\left(\mathbf{P}_{i+1} + Res((\mathbf{P}_{i+1}\oplus \mathbf{A}_{l,i+1}) *_1 \mathbf{k}_{i}^{'})\right),
\end{align}
where $\oplus$ represents the concatenation of feature maps. $\mathbf{P}_{i+1}$ denotes the features aggregated from the higher level $i+1$, and $\mathbf{P}_{3}$ is set to be $\mathbf{F}_{l,3}$. $\mathbf{k}_{i}^{'}$ denotes 3x3 convolution kernels for feature aggregation in level $i$. $Res(\cdot)$ denotes the residual blocks. $\phi_2$ is the bilinear interpolation with an upscaling factor of 2. Then, the output image $\mathbf{O}_l$ for the left view is given by projecting the feature maps to RGB space.

\subsection{Loss Function}

\mypara{Perceptual Loss.} Due to the possible tiny misalignment or intensity shift between the corrupted images and the clean images, measuring the similarity with the per-pixel penalty would bring ambiguity for pixels. Instead, we resort to perceptual loss~\cite{ChenKoltun2017}, which considers the similarity in feature space. Let $\{\Phi_t\}$ be a collection of layers in the pretrained VGG-16~\cite{simonyan2014very}, and each layer is a three-dimensional tensor. Here we use the `$relu1\_2$', `$relu2\_2$', `$relu3\_3$' and '$relu4\_3$' layers. Then the perceptual loss $L_P$ is defined as:
\begin{align}
    \sum_t\lambda_t(\|\Phi_t(\mathbf{I}_l) - \Phi_t(\mathbf{O}_l)\|_1 + \|\Phi_t(\mathbf{I}_r) - \Phi_t(\mathbf{O}_r)\|_1),
\end{align}
where $\|\cdot\|_1$ denotes the $l_1$ distance, and $\{\lambda_t\}$ are hyperparameters to balance the contribution of each layer $t$.

\mypara{Attention Consistency Loss.} To maintain the left-right consistency in output images, disparity maps estimated on clean stereo images through a pretrained stereo matching network are used as ground truths for supervision. Due to the inaccurate estimation by pretrained models, we experimentally find that blurry outputs are generated by directly warping each image to the other under the guidance of the estimated disparity map, followed by a distance measurement. Therefore, we constrain the consistency in feature space according to the ground-truth disparity maps and the attention scores in the RDA module.

The multiplication of the query $\mathbf{Q}$ and the key $\mathbf{K}$ gives the similarity between the feature grids in the left and right images. Therefore, we could expect that for each feature grid in the query view, the most similar feature grid in the key view can be found by applying the softmax function to $\mathbf{Q}^T\mathbf{K}$ along the line corresponding to the query feature grid. The difference between these two feature grids' indices is related to the disparity, and thus disparity maps $\mathbf{D}_l$ and $\mathbf{D}_r$ can be estimated. For simplicity, we use the left image as an example. The disparity estimated in each level $i$ is given by:
\begin{align}
  \mathbf{disp}_{i} = |\mathbf{P} - f(softmax(\mathbf{Q}_{l,i}^T\mathbf{K}_{r,i}))| , 
\end{align}
where $\mathbf{P}$ is the matrix containing the horizontal coordinates of each feature grid. $f$ is a function that computes the weighted summation of horizontal indices with the weight from attention scores, which is a differentiable alternative to $argmax$. The estimated disparity is then gradually merged with lower-level estimated disparity to get the final coarse disparity map:
\begin{align}
    \mathbf{D}_{i} = ((\phi_2(\mathbf{D}_{i+1}) \times 2)\oplus \mathbf{disp}_{i})*_1\mathbf{k}^{'},
\end{align}
where $\mathbf{k}^{'}$ denotes the 3x3 convolution kernels for merging the disparity maps from two levels. $\mathbf{D}_3$ is set to $\mathbf{disp}_3$. Following this rule, we can get a coarse disparity map.

Since left-right consistency only exists in non-occluded and non-border areas, masks $\mathbf{M}_l$ and $\mathbf{M}_r$ are generated along with disparity maps $\mathbf{\hat{D}}_l, \mathbf{\hat{D}}_r$ from GANet~\cite{zhang2019ga} to ensure the consistency in valid areas. The attention consistency loss is:
\begin{equation}
    L_C =\|(\mathbf{D}_{l} -  \mathbf{\hat{D}}_{l})\odot\mathbf{M}_l\|_1 + \|(\mathbf{D}_{r} -  \mathbf{\hat{D}}_{r})\odot\mathbf{M}_r\|_1.
\end{equation}


To enhance the consistency of disparities in different levels, the disparity map estimated from a higher level is upsampled and concatenated to the lower-level feature maps before applying the RDA module at a lower level.

\mypara{Full Objective.}
Our full objective is:
\begin{equation}
    L = L_P+\alpha L_C,
\end{equation}
where $\alpha$ represents the weight for $L_C$.

\section{Stereo Waterdrop Dataset}
Since there is no public stereo image dataset available for waterdrop removal, we collect a real-world dataset to benefit the research on stereo waterdrop removal. The dataset contains 837 stereo image pairs captured from 129 indoor and outdoor scenes with various waterdrops, disparities, and illumination conditions. We use the ZED 2 stereo camera for data collection. 

We mount a stereo camera and a piece of glass on two sturdy tripods, respectively. The angle between the stereo camera and the glass is randomly selected from 0 to 45 degrees. The distance between them varies from 2cm to 10cm to generate diverse waterdrop images. The two cameras share the same camera settings. For each scene, we first take one image pair with the clean glass to be the ground truth. Then several images are captured when we use droppers to randomly splash water on the glass, and other images are taken when using a sprayer to simulate more small, mist-like waterdrops. The ratio of the two types is 1:1. Around 6 to 8 image pairs with waterdrops are taken for each scene. In consideration of the refraction effect caused by the glass, the distance between the background scene and the glass, as well as the distance and the angle between the stereo camera and the glass, are kept fixed among all images captured in one scene. To minimize the influence of glass reflection, a black cloth is used to cover the back of the glass to block all transmissions from the back of the glass \cite{Lei2020}.


\section{Experiments}
\subsection{Experimental Setup}
We randomly split the stereo waterdrop dataset into training, validation, and test sets: 642 stereo image pairs from 100 scenes for training, 89 stereo pairs from 13 scenes for validation, and 106 stereo pairs from 16 scenes for testing. Images are resized to $624\times336$ for training. The stereo image pairs are randomly flipped horizontally and vertically for data augmentation.

The model is trained with batch size 3 for 70 epochs. We use an Adam optimizer~\cite{kingma2014adam} with the initial learning rate of 1e-4, and reduce the learning rate by a factor of 10 after 50 epochs. $\alpha$ is set to 5e-4. $\{\lambda_t\}$ are set to 1, 0.5, 0.4 and 1. 

\subsection{Comparison to State-of-the-art}
\mypara{Baselines.}
We compare our method with several state-of-the-art learning-based waterdrop removal approaches, including single-image methods by Eigen et al.~\cite{eigen2013restoring}, Isola et al.~\cite{isola2017image}, Qian et al.~\cite{qian2018attentive} and Quan et al.~\cite{quan2019deep}, and the multi-image method by Liu et al.~\cite{liu2020learning}. For single-image methods, we separate the image pair and test each image independently. The multi-image method requires 5 images from different viewpoints as input, and we reuse the image pair as reference images for fair comparison. We use the public pretrained models for all baselines except Pix2Pix~\cite{isola2017image} is trained from scratch and Qian et al.~\cite{qian2018attentive} is fine-tuned on our dataset because only the codes for the last two are publicly available.


\begin{table}[t!]
\centering 
\caption{Quantitative evaluation on the stereo waterdrop dataset.}
\setlength{\tabcolsep}{3mm}
\ra{1.15}
\begin{tabular}{@{}l@{\hspace{10mm}}c@{\hspace{7mm}}c@{\hspace{5mm}}c@{\hspace{5mm}}c@{}}
\toprule
&PSNR$\uparrow$   & MS-SSIM$\uparrow$ & LPIPS$\downarrow$ & Time(s)$\downarrow$ \\ 
\hline
Eigen et al.~\cite{eigen2013restoring} & 21.761 & 0.822 & 0.308 & 1.372\\
Pix2Pix~\cite{isola2017image} &22.758 &0.895 & 0.217 & \textbf{0.045}\\
Qian et al.~\cite{qian2018attentive} & 24.470  & 0.900& 0.163 & 0.083\\
Quan et al.~\cite{quan2019deep} & 24.972& 0.913 & 0.153 & 0.115\\
Liu et al.~\cite{liu2020learning} & 22.695 & 0.833  & 0.247 & 1.241\\
Ours & \textbf{26.064}  &\textbf{0.950} & \textbf{0.096} & 0.210\\
\bottomrule
\end{tabular}
\vspace{-2mm}
\label{tab:quantitative} 

\end{table}

\mypara{Quantitative Evaluation.}
To quantitatively evaluate our method, we use the standard PSNR, MS-SSIM~\cite{wang2003multiscale}, and LPIPS~\cite{zhang2018unreasonable} as the metrics. Although there might be a mild misalignment in the input-output image pairs (e.g., the moving clouds and the swaying trees), it exists in all methods, and thus the comparisons are fair. The inference time of a single image is evaluated on a single RTX 2080Ti.

The results are summarized in Table~\ref{tab:quantitative}. Our method achieves the best performance for all the metrics. This implies that stereo images provide additional useful information for waterdrop removal on both images, and more accurate correspondence can be found through row-wise dilated attention and attention consistency loss.

\begin{table}[t!]
\centering
\caption{Perceptual experiments.}
\setlength{\tabcolsep}{3mm}
\ra{1.15}
\begin{tabular}{@{}lc@{}}
\toprule
& Preference rate \\
\midrule
Ours$>$Eigen et al.~\cite{eigen2013restoring}& 99.90\%\\
Ours$>$Pix2Pix~\cite{isola2017image}& 99.62\%\\
Ours$>$Qian et al.~\cite{qian2018attentive} & 100.00\%\\
Ours$>$Quan et al.~\cite{quan2019deep} & 99.53\%\\
Ours$>$Liu et al.~\cite{liu2020learning}  & 100.00\% \\
\bottomrule
\end{tabular}
\vspace{-7mm}

\label{tab:user_study}
\end{table}

\begin{figure*}[!ht]
    \centering 
    \begin{tabular}{@{\hspace{0.1mm}}c@{\hspace{0.7mm}}c@{\hspace{0.7mm}}c@{\hspace{0.7mm}}c@{\hspace{0.7mm}}c@{}}
      &Left & Right & Left & Right \\
     \rotatebox{90}{\hspace{7mm}Input} &
     \includegraphics[width=0.24\linewidth]{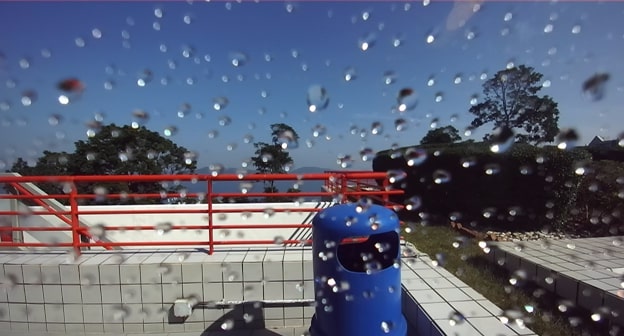} &    \includegraphics[width=0.24\linewidth]{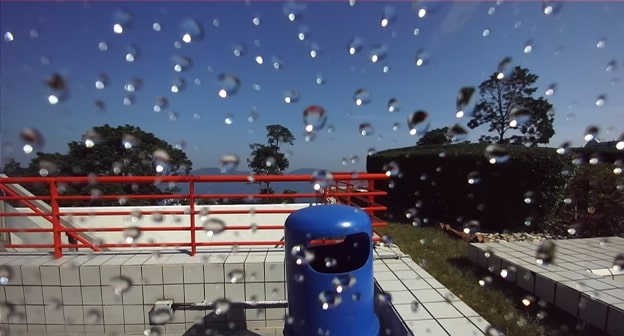}  & \includegraphics[width=0.24\linewidth]{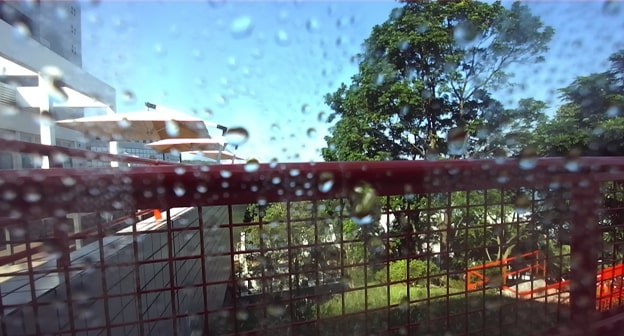} &    \includegraphics[width=0.24\linewidth]{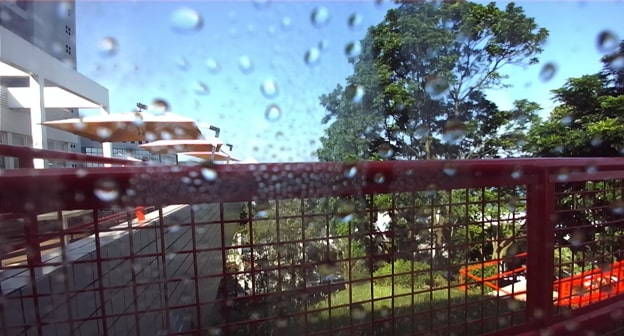}  \\ 
    
    \rotatebox{90}{\hspace{10mm}GT} &
     \includegraphics[width=0.24\linewidth]{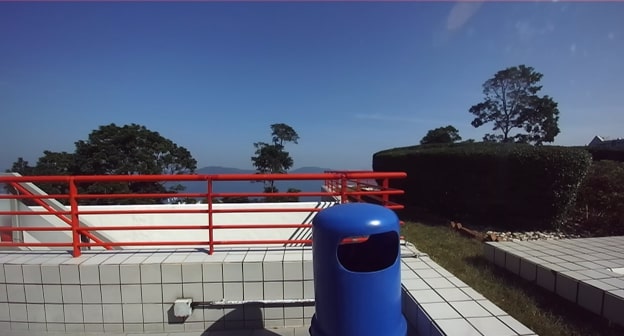} &    \includegraphics[width=0.24\linewidth]{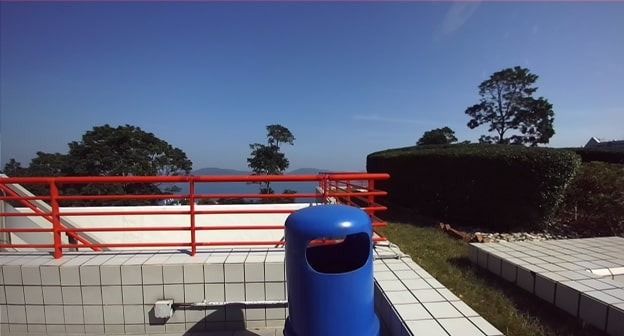} & \includegraphics[width=0.24\linewidth]{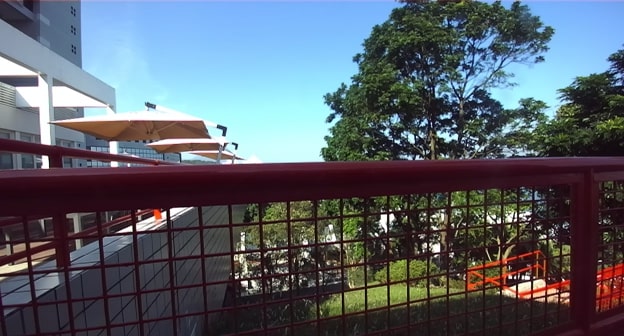} &    \includegraphics[width=0.24\linewidth]{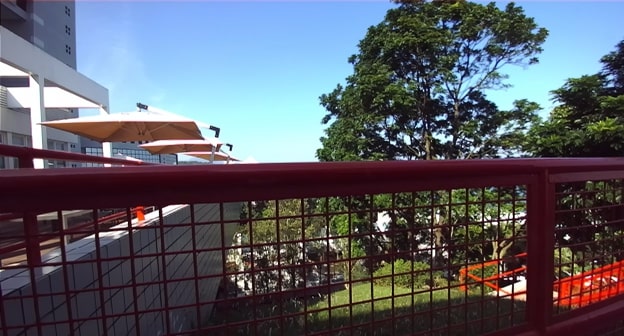}  \\
     
     \rotatebox{90}{\hspace{9mm}~\cite{eigen2013restoring}} &
     \includegraphics[width=0.24\linewidth]{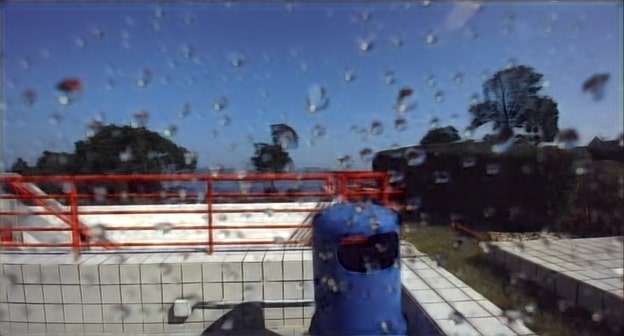} &    \includegraphics[width=0.24\linewidth]{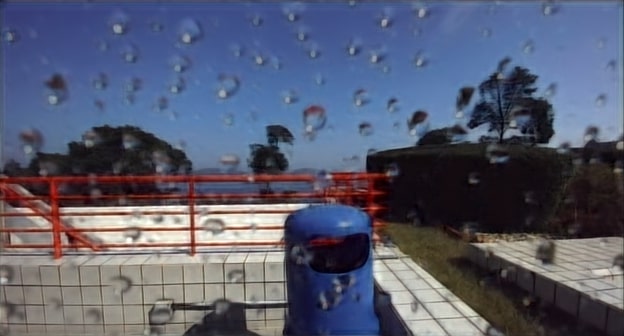}  & \includegraphics[width=0.24\linewidth]{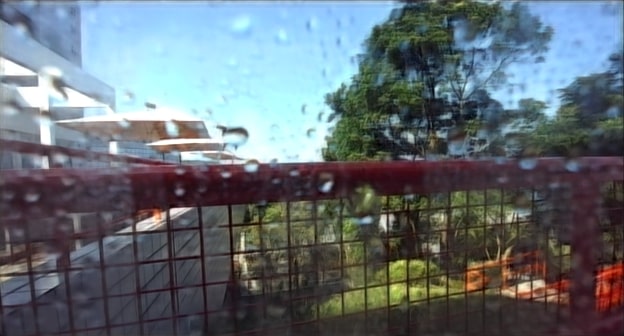} &    \includegraphics[width=0.24\linewidth]{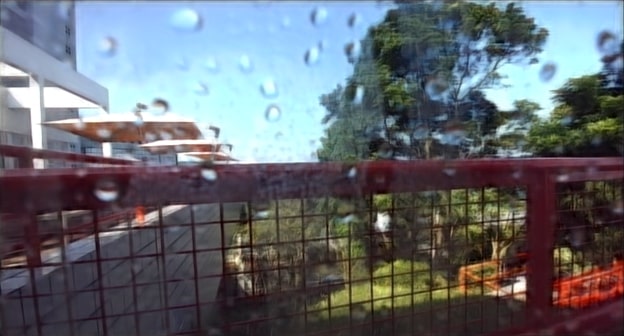}   \\ 
     
     \rotatebox{90}{\hspace{8mm}~\cite{isola2017image}} &
     \includegraphics[width=0.24\linewidth]{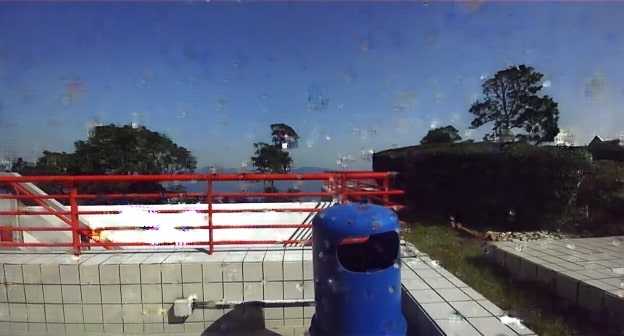} &    \includegraphics[width=0.24\linewidth]{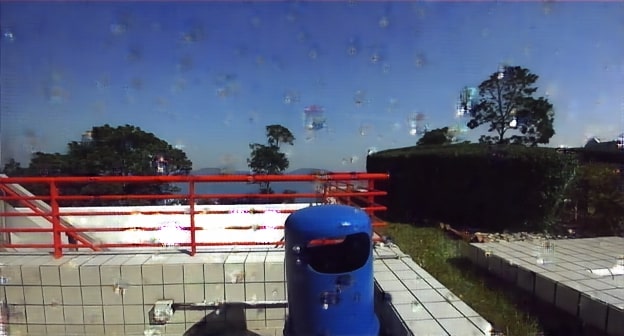}  & \includegraphics[width=0.24\linewidth]{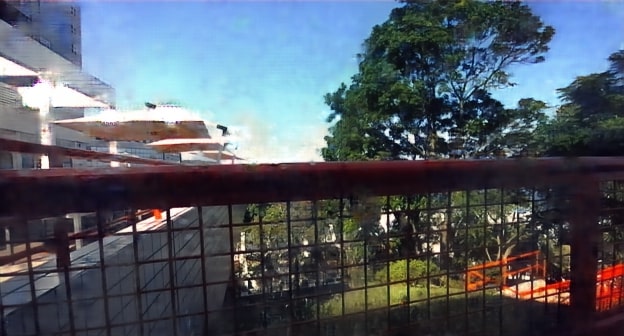} &    \includegraphics[width=0.24\linewidth]{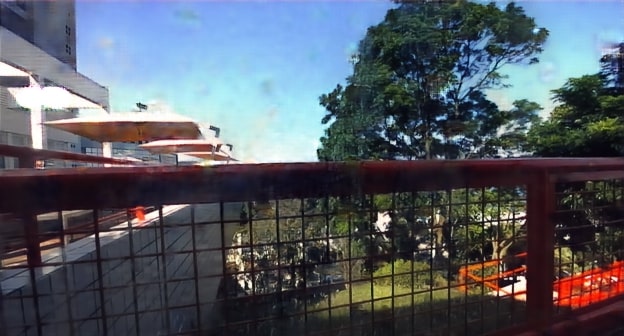}   \\ 
    \rotatebox{90}{\hspace{8mm}~\cite{qian2018attentive}} &
     \includegraphics[width=0.24\linewidth]{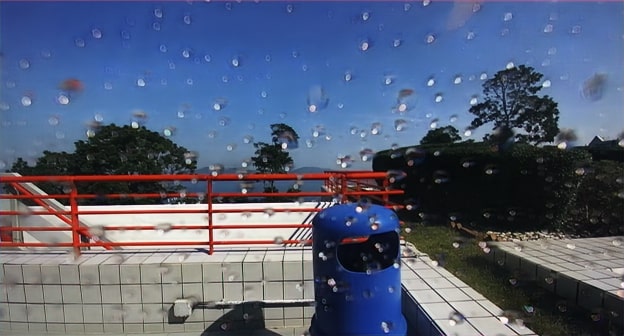} &    \includegraphics[width=0.24\linewidth]{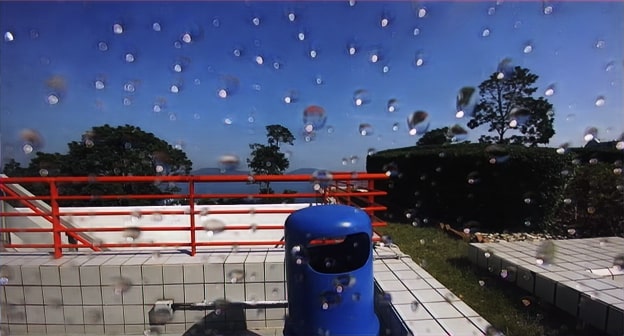}  & \includegraphics[width=0.24\linewidth]{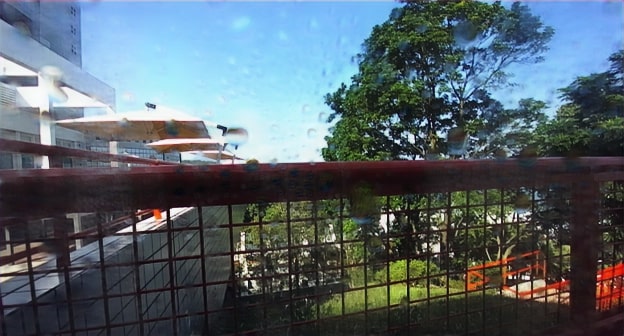} &    \includegraphics[width=0.24\linewidth]{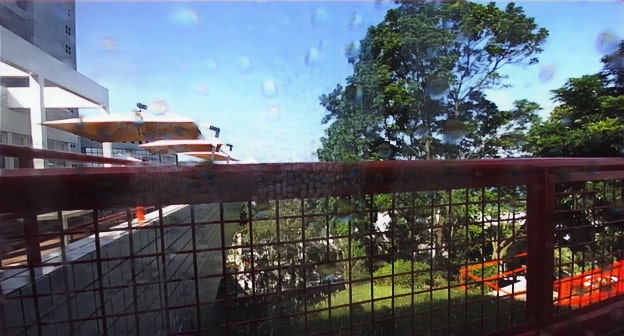}   \\ 
    \rotatebox{90}{\hspace{8mm}~\cite{quan2019deep}} &
     \includegraphics[width=0.24\linewidth]{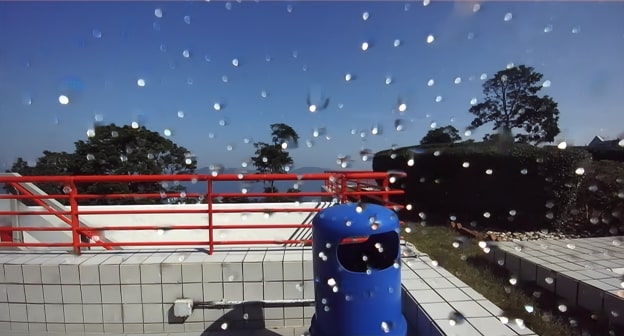} &    \includegraphics[width=0.24\linewidth]{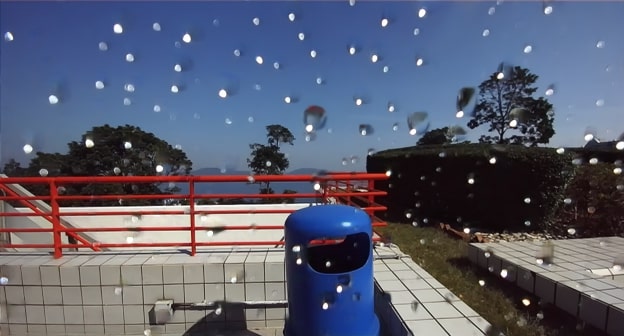}  & \includegraphics[width=0.24\linewidth]{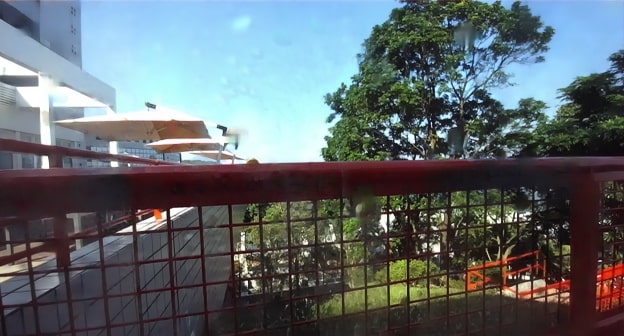} &    \includegraphics[width=0.24\linewidth]{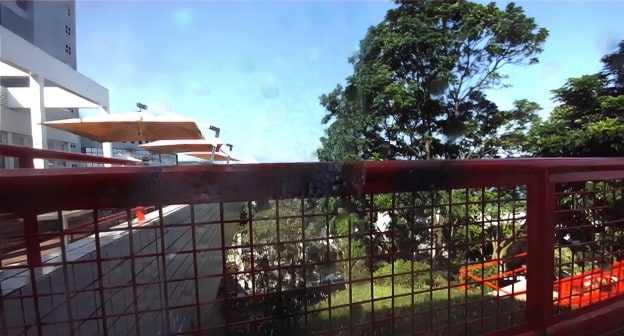}   \\ 
     
     \rotatebox{90}{\hspace{8mm}~\cite{liu2020learning} } &
     \includegraphics[width=0.24\linewidth]{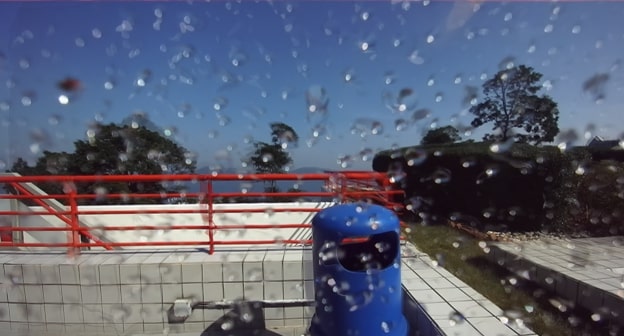} &    \includegraphics[width=0.24\linewidth]{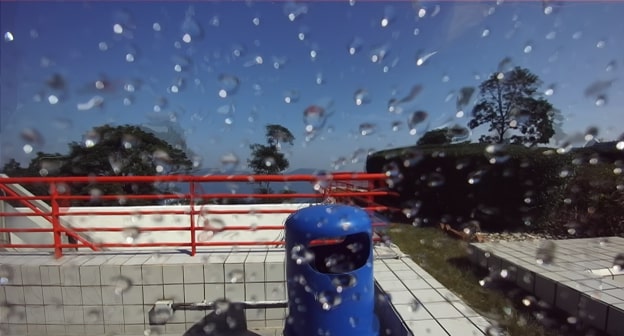}  & \includegraphics[width=0.24\linewidth]{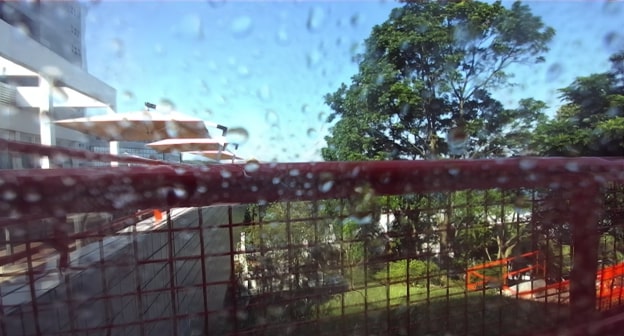} &    \includegraphics[width=0.24\linewidth]{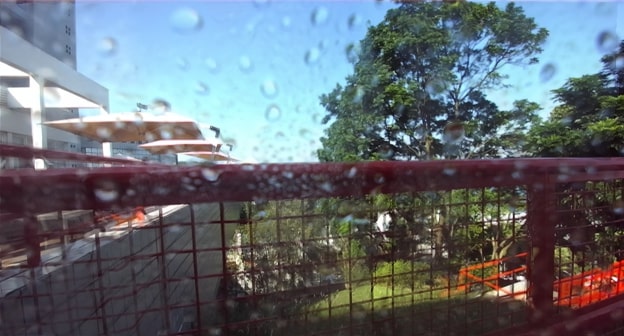}    \\

    \rotatebox{90}{\hspace{7mm}Ours} &
      \includegraphics[width=0.24\linewidth]{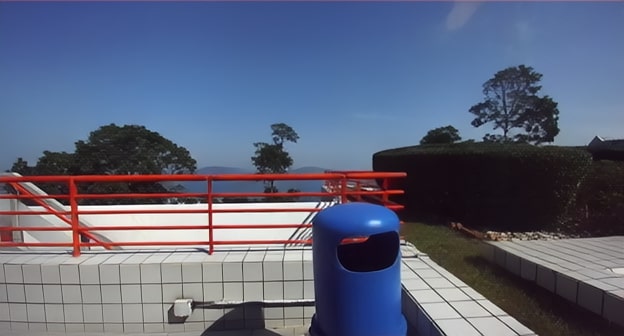} &    \includegraphics[width=0.24\linewidth]{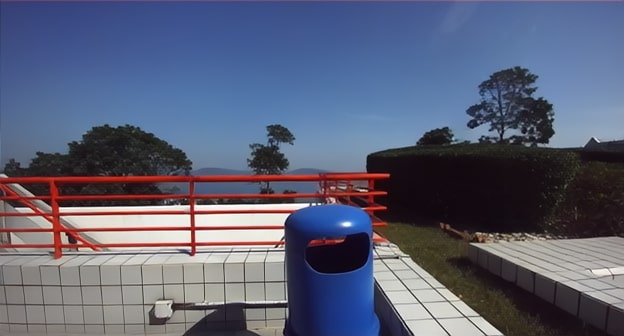}  & \includegraphics[width=0.24\linewidth]{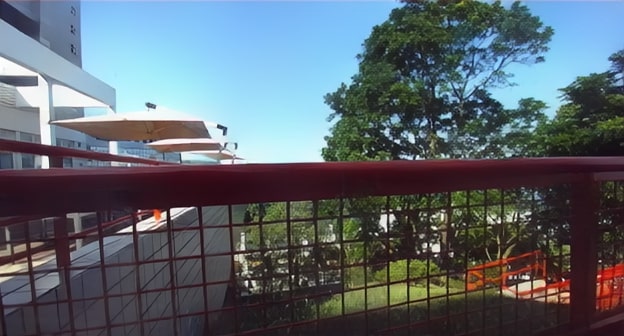} &    \includegraphics[width=0.24\linewidth]{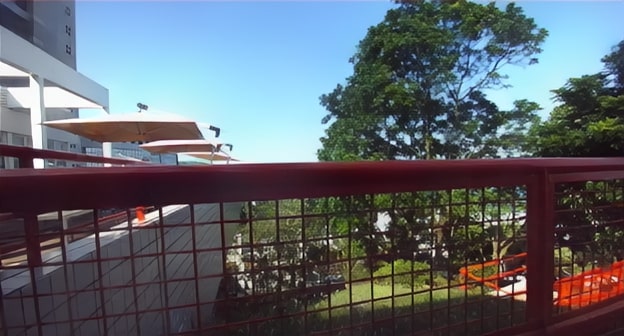}  
     
    \end{tabular}
    \vspace{-2mm}
    \caption{Qualitative results. Waterdrops are removed and details are recovered by our method compared with baseline methods.}
    \label{ fig:baseline}
    \vspace{-4mm}
\end{figure*}

\mypara{Qualitative Results.}
Figure~\ref{ fig:baseline} shows the qualitative comparisons against baseline methods. The results on images with sparse and dense waterdrops are reported. The baseline methods find it hard to handle the case with various colors in one waterdrop caused by the illumination or background scene, as shown in the first example. They also fail to remove all the small and dense waterdrops, as shown in the second example. Even if they can detect the waterdrops correctly, some of the obstructed areas are filled with inconsistent textures. In contrast, our method can successfully remove waterdrops of all sizes and leverage the stereo information to reconstruct the fine details.

\mypara{Perceptual Experiments.}
To evaluate the perceptual quality of the output images, we conduct a perceptual experiment on Amazon Mechanical Turk. In each task, a randomized A/B test is conducted against a baseline method on all the test images. The participants are presented with two images from the same viewpoint each time in random order: a result from one baseline method and another from ours. We ask them to ``choose the clearer image with fewer waterdrops". A total number of 50 workers were involved in the perceptual experiments. The results are shown in Table~\ref{tab:user_study}. Our method is preferred over baseline methods by a large margin.

\subsection{Controlled Experiments}
We conduct controlled experiments to analyze our method. All the models are retrained on the stereo waterdrop dataset with hyperparameter tuning for fair comparisons. 

\begin{table}[!t]
\centering 
\caption{Controlled experiments on the stereo waterdrop dataset.}
\setlength{\tabcolsep}{3mm}
\ra{1.2}
\begin{tabular}{@{}l@{\hspace{5mm}}c@{\hspace{5mm}}c@{\hspace{5mm}}c@{}} 
\toprule
& PSNR$\uparrow$ & MS-SSIM$\uparrow$  & LPIPS$\downarrow$   \\ 
\hline
Ours-PASSRnet~\cite{wang2019learning}  & 24.175 & 0.917 & 0.128\\
Ours-AANet~\cite{xu2020aanet}  &  24.673 & 0.920  & 0.125 \\
\hline
Ours-mono & 25.027 & 0.934 & 0.117\\
\hline
Ours-TTT & 25.154 & 0.938 & 0.112 \\
Ours-RTT & 25.708 & 0.946 & 0.104
\\
Ours-RRR & 26.061 & 0.950 & 0.098 \\
Ours-FD & 25.603 & 0.943 & 0.106 \\
\hline
Ours-1row & 25.837 & 0.944 & 0.105 \\
Ours-5row & 25.633 & 0.942 & 0.107 \\
\hline
Ours-nocat & 25.986 & 0.945 & 0.099 \\
Ours-noAC & 25.851 & 0.942 & 0.103\\
\hline
Ours & \textbf{26.064}  &\textbf{0.950} & \textbf{0.096}\\
\bottomrule
\end{tabular}
\vspace{-4mm}
\label{tab:ablation} 

\end{table}

\mypara{Ours vs. Other Consistency Strategies.}
We model the stereo consistency through RDA module and attention consistency loss. There exist some alternatives to ensure consistency, such as parallax attention and cost volume. We incorporate two structures into our framework to enforce consistency: PASSRnet~\cite{wang2019learning} that adopts parallax attention to maintain the consistency in stereo super-resolution task, and AANet~\cite{xu2020aanet} which utilizes cost volumes to estimate the consistency for stereo matching. For PASSRnet, we replace the RDA module with the parallax attention module in PASSRnet (``Ours-PASSRnet"). We keep the losses used in~\cite{wang2019learning} unchanged, except that SR loss is replaced by perceptual loss, and photometric loss is removed since per-pixel consistency does not hold in the input pair. For AANet, we warp the reference image with the disparity map estimated by AANet (``Ours-AANet"). The warped image is then concatenated with the query image and put into the refinement module. The network outputs estimated disparity map and restored image, and $l_1$ loss and perceptual loss are used for training. Results in Table~\ref{tab:ablation} show that these two methods suffer from performance degradation in all the metrics because of the limited receptive fields in parallax attention and cost volume.

\mypara{Single Input vs. Stereo Input.} Compared to a single image, stereo images provide additional information from a different viewpoint. To verify the benefit of stereo images, we change the input pair with two identical images from a single view (``Ours-mono") and retrain the network with the perceptual loss $L_P$ only. The decrease of PSNR and MS-SSIM and the increase of LPIPS score in Table~\ref{tab:ablation} show the significance of using the image from a different view to detect waterdrops robustly and bring back the details. 


\mypara{Row-wise Dilated Attention vs. Typical Attention.} To demonstrate the effectiveness of the RDA module, we replace RDA with typical non-local attention, where only 1x1 convolution kernels are used. Results are shown in Table~\ref{tab:ablation}. The model with typical attention modules for all levels (``Ours-TTT") suffers from performance degradation. If RDA in the second level is removed (``Ours-RTT"), the performance deteriorates as well, but it is better than the one without RDA. They show the necessity of a larger receptive field in low-level features. If RDA is used in all the modules (``Ours-RRR"), the performance is comparable to the one without RDA in the third level (``Ours"). Considering the extra computational cost in RDA, we choose the typical attention module for the third level. Moreover, we verify the unnecessary dilation for the value in the RDA module (``Ours-FD"). The feature maps convolve with the dilated convolution kernels with factors 1, 2, and 4 to generate the value. The performance becomes worse as a result of the unnecessary aggregation of the surrounding features.


\mypara{Number of Rows.} Considering the epipolar constraint in stereo images, attention is conducted on several rows instead of the whole feature map. We select different numbers of rows to train: 1-row attention with stride 1, 5-row attention with stride 2, and 3-row attention with stride 2 (``Ours"). From Table~\ref{tab:ablation}, we find that both small and large numbers degrade the performance. With small number of rows, the key and value are constrained on a single line which is unfriendly for occluded areas that would like to aggregate features from lines above or below. A large number results in intensive computational cost, and the model is prone to average the critical features with the surroundings.

\mypara{Losses.} We adopt two losses: perceptual loss and attention consistency loss. Perceptual loss is necessary as it ensures the visual quality. To test the effectiveness of attention consistency loss, we train our model with the perceptual loss only (``Ours-noAC"). The performance without attention consistency loss gets worse, as shown in Table~\ref{tab:ablation}. We also report the result without the estimated disparity map concatenated to the feature maps (``Ours-nocat"). The result is not as good as the one with disparity concatenation.

\begin{figure*}[!h]
    \centering 
    \begin{tabular}{@{\hspace{1mm}}c@{\hspace{1mm}}c@{\hspace{1mm}}c@{\hspace{1mm}}c@{\hspace{1mm}}c@{\hspace{1mm}}c@{\hspace{1mm}}c@{\hspace{1mm}}c@{\hspace{1mm}}c@{}}
    & Input & \cite{eigen2013restoring} & \cite{isola2017image} &\cite{qian2018attentive} &\cite{quan2019deep}& \cite{liu2020learning} & Ours &GT\\
    \rotatebox{90}{\hspace{-4mm} Left} &
     \raisebox{-.5\height}{\includegraphics[width=0.115\linewidth]{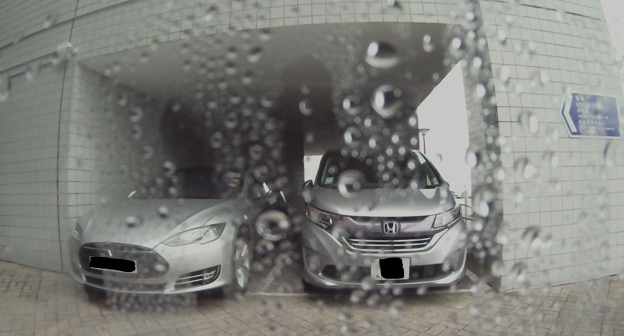}} &    \raisebox{-.5\height}{\includegraphics[width=0.115\linewidth]{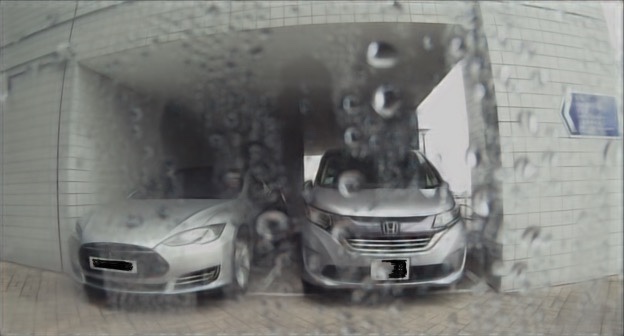}}  &
     \raisebox{-.5\height}{\includegraphics[width=0.115\linewidth]{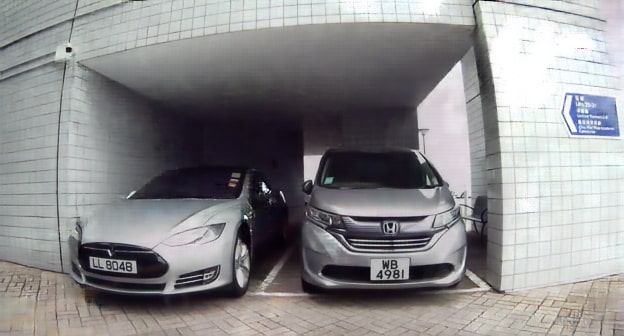}}  &
     \raisebox{-.5\height}{\includegraphics[width=0.115\linewidth]{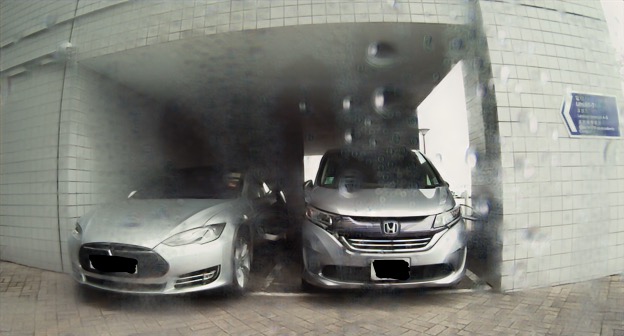}}  &
     \raisebox{-.5\height}{\includegraphics[width=0.115\linewidth]{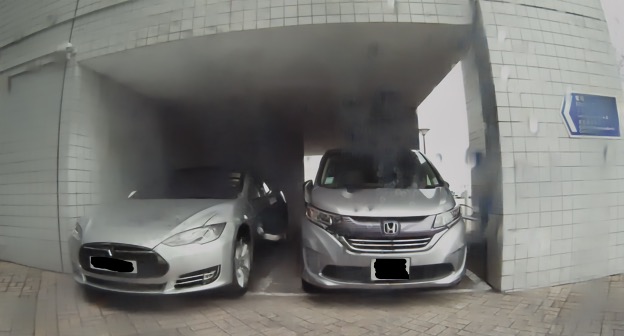}}  &
     \raisebox{-.5\height}{\includegraphics[width=0.115\linewidth]{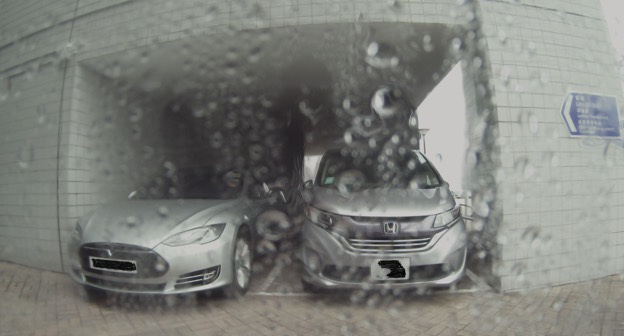}}  &
     \raisebox{-.5\height}{\includegraphics[width=0.115\linewidth]{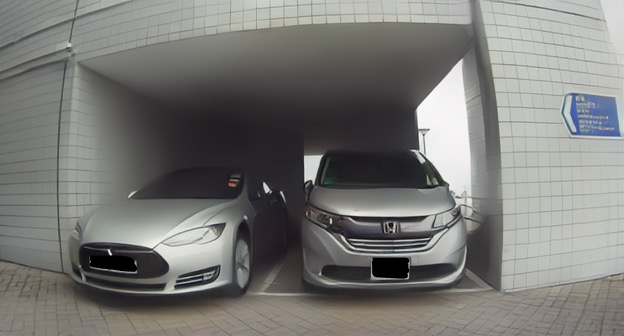}} &
     \raisebox{-.5\height}{\includegraphics[width=0.115\linewidth]{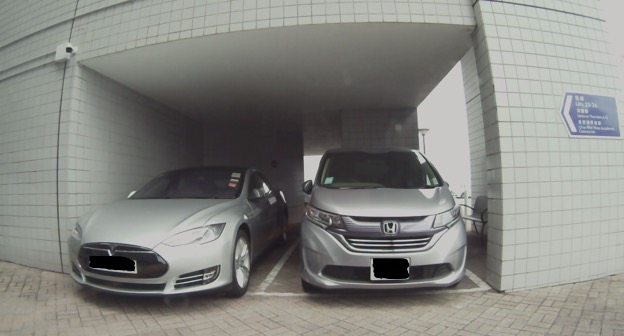}}\\
     \rotatebox{90}{\hspace{-5mm} Right} &
     \raisebox{-.5\height}{\includegraphics[width=0.115\linewidth]{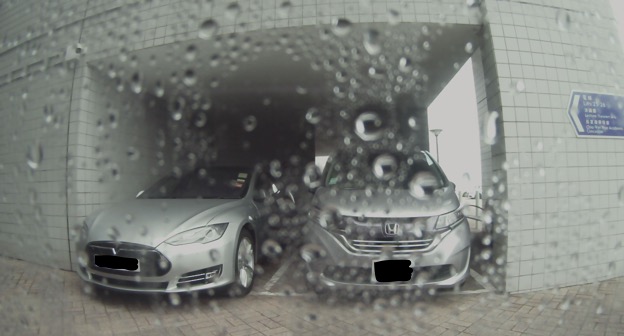}} &    \raisebox{-.5\height}{\includegraphics[width=0.115\linewidth]{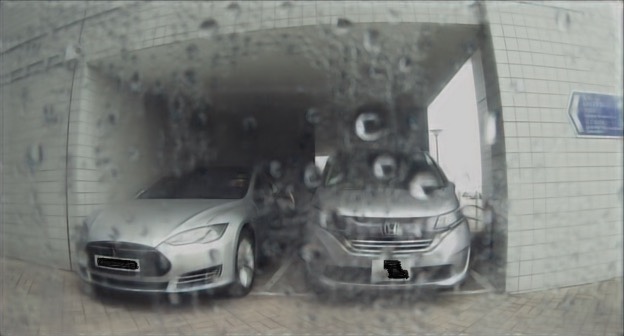}}  &
     \raisebox{-.5\height}{\includegraphics[width=0.115\linewidth]{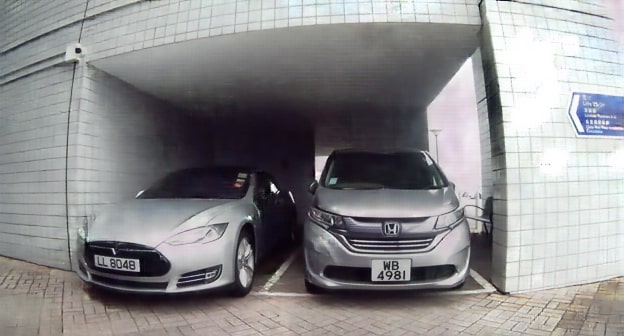}}  &
     \raisebox{-.5\height}{\includegraphics[width=0.115\linewidth]{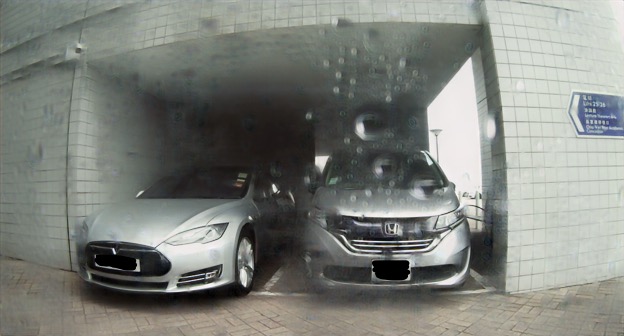}}  &
     \raisebox{-.5\height}{\includegraphics[width=0.115\linewidth]{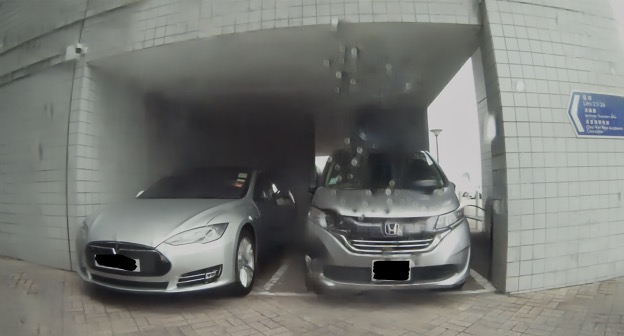}} &
     \raisebox{-.5\height}{\includegraphics[width=0.115\linewidth]{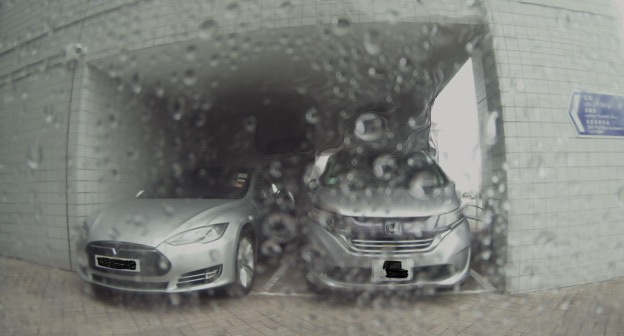}}  &
     \raisebox{-.5\height}{\includegraphics[width=0.115\linewidth]{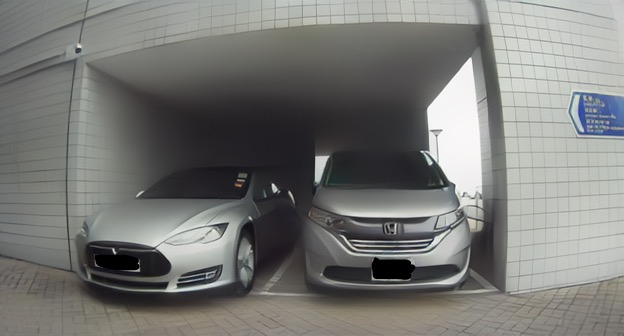}}& \raisebox{-.5\height}{\includegraphics[width=0.115\linewidth]{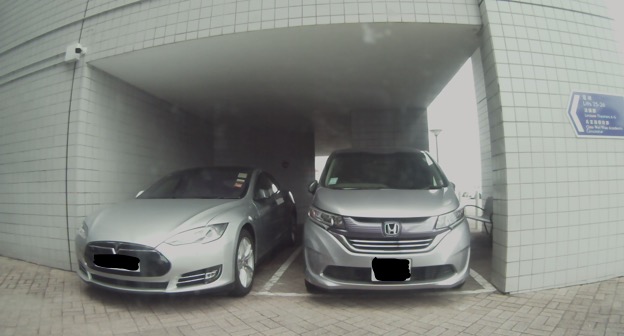}}\\
     \rotatebox{90}{\hspace{-3mm} Left} &
     \raisebox{-.5\height}{\includegraphics[width=0.115\linewidth]{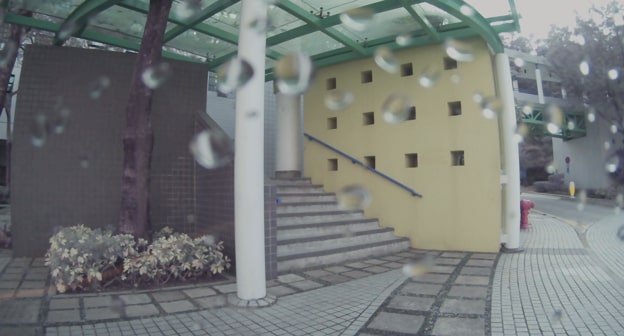}} &    \raisebox{-.5\height}{\includegraphics[width=0.115\linewidth]{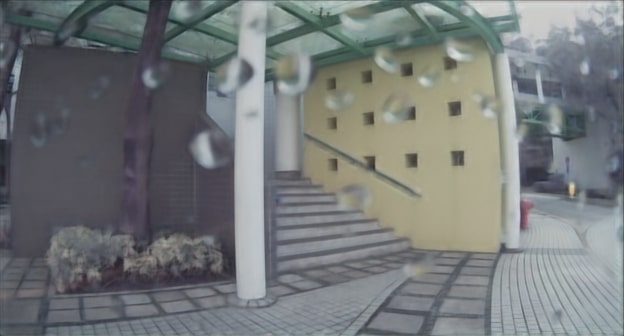}}  &
     \raisebox{-.5\height}{\includegraphics[width=0.115\linewidth]{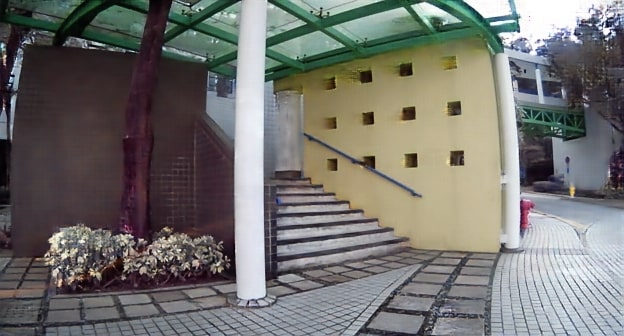}}  &
     \raisebox{-.5\height}{\includegraphics[width=0.115\linewidth]{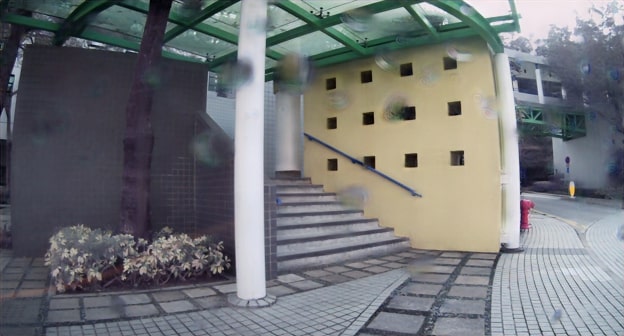}}  &
     \raisebox{-.5\height}{\includegraphics[width=0.115\linewidth]{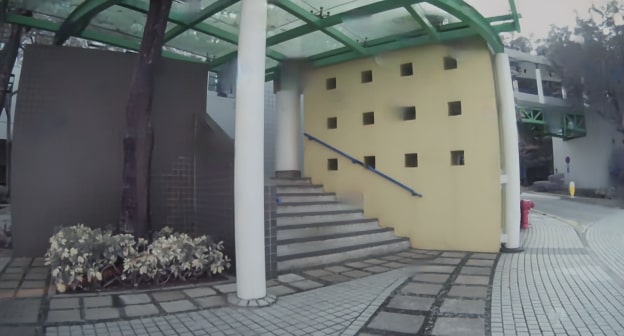}}&
     \raisebox{-.5\height}{\includegraphics[width=0.115\linewidth]{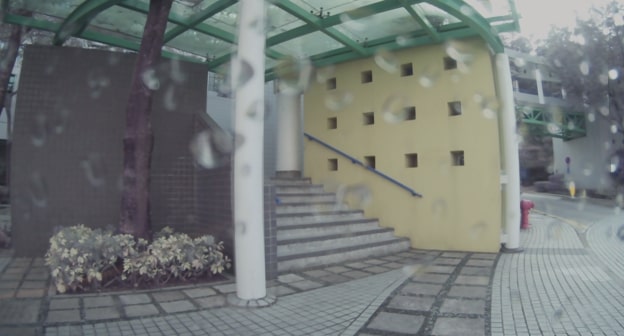}}  &
     \raisebox{-.5\height}{\includegraphics[width=0.115\linewidth]{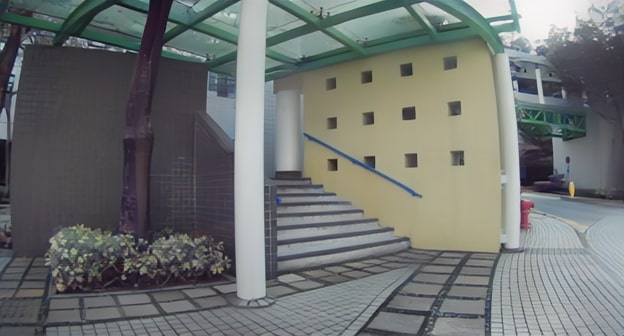}} & \raisebox{-.5\height}{\includegraphics[width=0.115\linewidth]{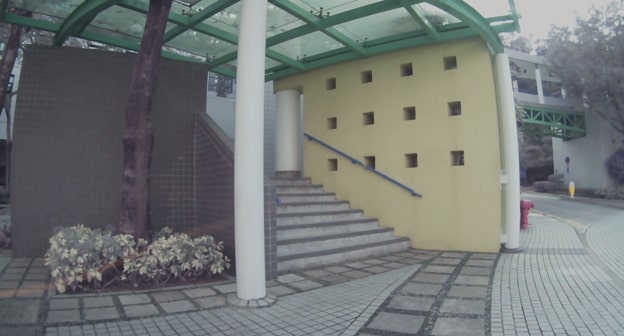}} \\
     \rotatebox{90}{\hspace{-3mm}Right} &
     \raisebox{-.5\height}{\includegraphics[width=0.115\linewidth]{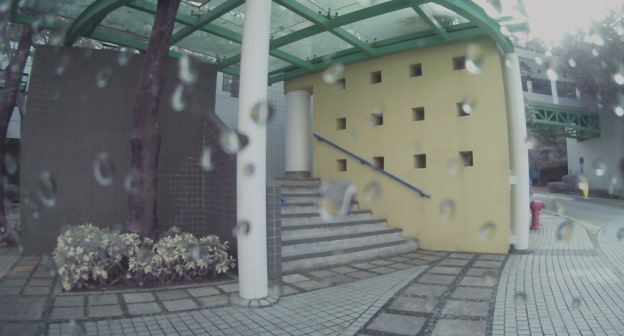}} &    \raisebox{-.5\height}{\includegraphics[width=0.115\linewidth]{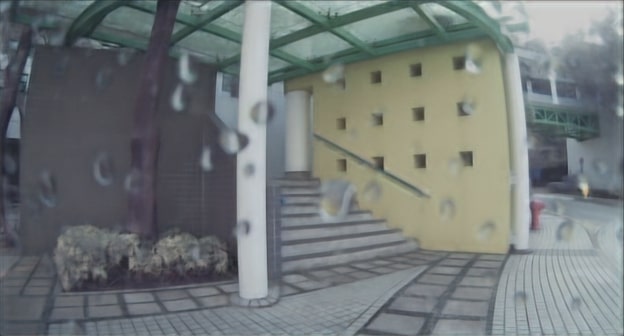}}  &
     \raisebox{-.5\height}{\includegraphics[width=0.115\linewidth]{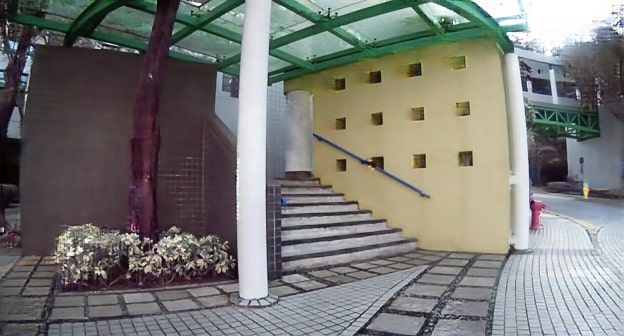}}  &
     \raisebox{-.5\height}{\includegraphics[width=0.115\linewidth]{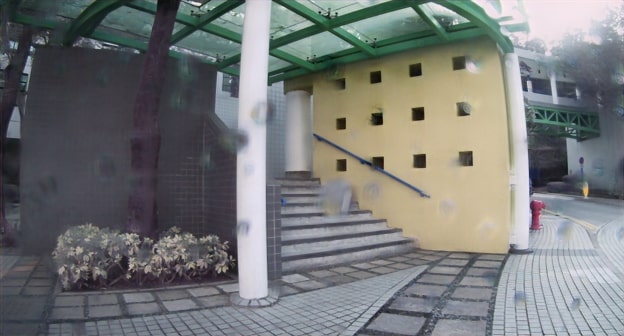}}  &
     \raisebox{-.5\height}{\includegraphics[width=0.115\linewidth]{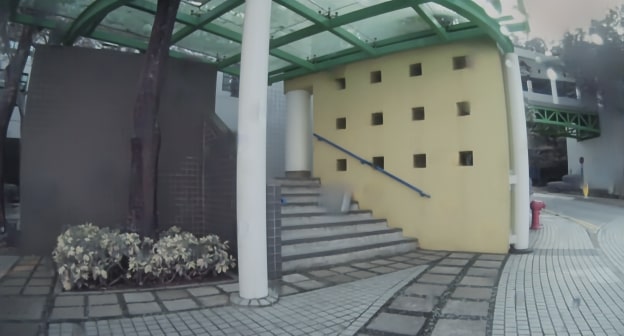}}  &
     \raisebox{-.5\height}{\includegraphics[width=0.115\linewidth]{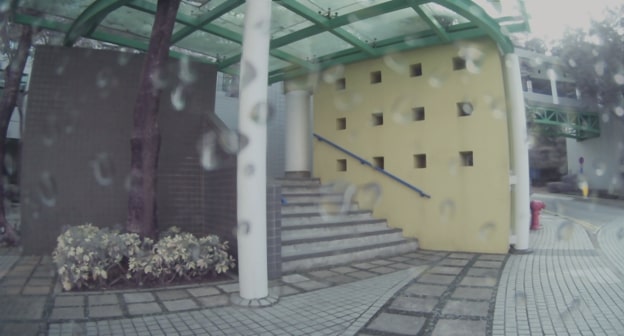}}  &
     \raisebox{-.5\height}{\includegraphics[width=0.115\linewidth]{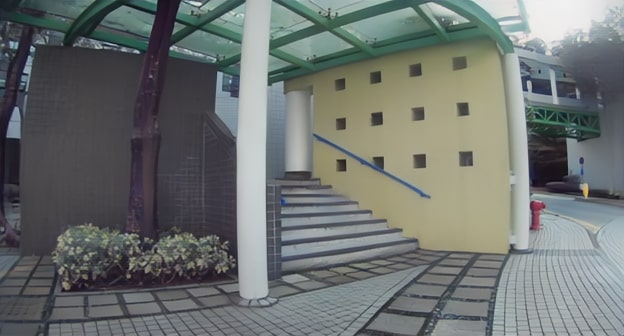}}& \raisebox{-.5\height}{\includegraphics[width=0.115\linewidth]{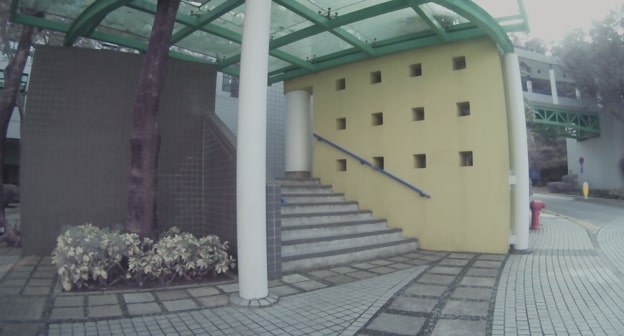}}
    \end{tabular}
    \caption{Generalization ability. Ours is trained on ZED 2 dataset and tested on images of MYNT EYE camera. Zoom in for details.}
    \vspace{-4mm}
    \label{fig:generalize}
\end{figure*}

\subsection{Generalization}
To verify the generalization ability of our proposed method, we use the model trained on ZED 2 dataset to test on images captured by the MYNT EYE camera. The results are shown in Fig.~\ref{fig:generalize}. Ours still outputs cleaner and detail-preserving images compared to state-of-the-art methods.
\section{Conclusion}
We have presented a learning-based approach for stereo waterdrop removal, where row-wise dilated attention is proposed to enlarge attention's receptive field for better left-right information propagation in corrupted stereo images, and attention consistency loss further enhances the consistency in the stereo image pair. To evaluate different methods on stereo real data, we collect a real-world stereo dataset for waterdrop removal. The experiments have demonstrated that our approach achieves excellent performance on waterdrop removal with stereo images, as indicated in the user study. We hope our work can inspire researchers to explore other image enhancement tasks with stereo images in the future.

{\small
	\bibliographystyle{ieee_fullname} %
	\bibliography{reference}
}

\end{document}